\documentclass{article}

\usepackage{arxiv}

\usepackage[utf8]{inputenc} 
\usepackage[T1]{fontenc}    
\usepackage{hyperref}       
\usepackage{url}            
\usepackage{booktabs}       
\usepackage{amsfonts}       
\usepackage{nicefrac}       
\usepackage{microtype}      
\usepackage{lipsum}
\usepackage{graphicx}
\usepackage{framed}
\usepackage{setspace}
\usepackage{caption}
\usepackage{enumitem}
\graphicspath{ {./images/} }

\title{What does it mean to understand language?}

\author{
 Colton Casto \\
  Kempner Institute for the\\Study of Natural \& Artificial Intelligence\\
  Harvard University\\
  Cambridge, MA 02138 \\
  \texttt{ccasto@mit.edu} \\
   \And
 Anna Ivanova \\
  School of Psychology\\
  Georgia Institute of Technology\\
  Atlanta, GA 30332 \\
  \texttt{a.ivanova@gatech.edu} \\
  \And
 Evelina Fedorenko \\
  Department of Brain and Cognitive Sciences\\
  Massachusetts Institute of Technology\\
  Cambridge, MA 02139 \\
  \texttt{evelina9@mit.edu} \\
  \And
 Nancy Kanwisher \\
  Department of Brain and Cognitive Sciences\\
  Massachusetts Institute of Technology\\
  Cambridge, MA 02139 \\
  \texttt{ngk@mit.edu} \\
}

\begin{document}
\maketitle 
\begin{abstract}
\begin{spacing}{1.25}
Language understanding entails not just extracting the surface-level meaning of the linguistic input, but constructing rich mental models of the situation it describes. Here we propose that because processing within the brain’s core language system is fundamentally limited, deeply understanding language requires exporting information from the language system to other brain regions that construct mental models of minds, objects, and places, and that compute perceptual and motor representations. Building on earlier theorizing and empirical work, we argue that recent progress in cognitive neuroscience provides both the conceptual foundation and the methods to directly evaluate this hypothesis, thus opening up a new strategy to reveal what it means, cognitively and neurally, to understand language.
\end{spacing}
\end{abstract}


\newpage
\section*{A cognitive neuroscience perspective on language understanding}
What does it mean to understand language? Let's consider a few sentences:  
\begin{itemize}[itemsep=1.5pt]
\itshape
  \item "Killer whales exfoliate each other with home-made scrubbers."
  \item "Democracies facing authoritarian threats push back with mass demonstrations."
  \item "The man transfers the wet glue from one piece of wood to another."
  \item "More people have been to Russia than I have."
\end{itemize}

\begin{spacing}{1.25}
Did you visually imagine the killer whales? (What were those scrubbers made of?)  \\
Did you relate the authoritarian threats to a particular democracy (perhaps one close to home)?  \\
Did you wonder exactly how the glue got transferred? (by scraping the glue off one piece of wood with a knife then smearing it on the other, or by pressing the two pieces together?)  \\
Did you realize that the last sentence doesn't make any sense at all [1]? 

What then does it mean to understand language, cognitively, computationally, and neurally? How rich and structured is your mental representation of the sentence you are reading right now, and what cognitive and neural systems are holding this representation? And are you now linking the meaning of this sentence to the previous one, to the unfolding structure of the argument in this article, and to thoughts you have had previously about language processing? In this opinion piece, we argue that a \textbf{deep understanding} of language, which entails building mental models and rich representations of meaning that connect to our broader knowledge of the world (see \textit{\textbf{Glossary}}), requires the \textbf{exportation} of information from the brain’s core language system to various perceptual, motor, and cognitive systems that can build models of what we are hearing or reading.

The idea that language can be processed at multiple levels is not new. Decades of theorizing and behavioral experimentation have supported the idea that, in addition to parsing the surface form of sentences and building gist-like representations of their meaning, we construct \textbf{situation models} that diagram in our minds the characters described, their goals and spatial locations, and the temporal and causal relations among the events taking place [2–9]. Other frameworks have similarly emphasized the difference between linguistic processing and deeper simulation-based or conceptual processing (e.g., [10,11]), connecting to longstanding debates in cognitive psychology about the format and organization of concepts [12–19]. Although our proposal builds upon these earlier ideas, it extends and refines them in several respects (\textbf{Box 1}).

In the decades since these early proposals, cognitive neuroscience has provided both a rich taxonomy of mental processes that selectively engage specific brain regions, and the methods to directly measure their recruitment during language understanding. In particular, neuroscientists have leveraged fMRI and other approaches to identify and functionally characterize both the \textbf{core language system} and many other perceptual, motor, and cognitive systems that it may export information to for further processing (\textbf{Figure 1A}). A key finding from this research is that the core language system is specifically engaged in understanding (and producing) language, and not in other nonlinguistic cognitive processes (see [20] for review). Further, many other cortical regions show remarkable specificity for other mental functions, including regions in the ventral visual pathway selectively engaged in perception of faces, bodies, and scenes [21–23], regions of auditory cortex selective for the perception of speech sounds and music [24,25], the right temporo-parietal junction (rTPJ) [26] and the other regions comprising the theory of mind network [27,28] which are selectively engaged in thinking about mental states, and multiple regions specifically engaged in visually-guided navigation [29] and in reaching and grasping [30]. Ongoing research is investigating the precise function of many other brain regions, including those engaged in intuitive physical reasoning [31], the perception of social agents and their actions [32], amodal semantic processing [33], episodic retrieval and self-projection [27,34], and aspects of formal reasoning [35,36]. 
\end{spacing}
\begin{framed}
\textbf{Box 1: Relationship to debates about embodied language processing.}

Our proposal builds upon prior work—rooted in classic studies on the nature of conceptual representations in the mind and brain [12–19]—arguing that language understanding entails activation of sensorimotor or other “embodied” representations [10,11,85]. We share with some of these proposals (e.g., [10,11]) both an appreciation of the richness of the semantic information captured by language statistics alone, and a recognition that human language understanding requires far more than these linguistic approximations of meaning.

However, our exportation hypothesis is motivated by specific findings from cognitive neuroscience. In particular, the critical foundation for this proposal consists of the recent discoveries that i) the brain’s language system extracts relatively shallow and language-specific meanings, and ii) that there exist a number of functionally-specific brain systems that build mental models of minds [26], objects [31], and places [22,28]. This foundation allows us to extend and refine early theories in two ways. First, the existence of certain functionally-specific brain regions allows for new predictions about the kinds of meanings that could be represented, in part, outside the language system. This includes meanings related to other minds, the physical world, and places, broadening prior proposals, which have emphasized the relationship between linguistic and sensorimotor (plus internal metacognitive) representations [12,14,16]. Second, the hypothesis that the language system builds shallow meanings makes predictions about how the responsibility of constructing meaning could be divided among the core language system and the non-linguistic cognitive, perceptual, and motor brain systems. Specifically, it suggests that an approximate representation of meaning is initially constructed by the language system—derived from our knowledge of language statistics [42,49]—that is later elaborated on by non-linguistic brain systems. This idea refines prior proposals (e.g., [10,11]), which have differed in the role they ascribe to the language system in building meaning. 

Equally important, our proposal makes concrete suggestions for how to test these ideas \textit{empirically}. Although some prior studies have found support for the idea that a linguistic input can activate sensorimotor—and, to some extent, emotional, cognitive, and metacognitive—representations (e.g., [69,70,79,116–118]), the empirical record is far from settled. Some of the effects that initially supported embodied language understanding have since been shown to be strongly task- or context-dependent (e.g. [71,84,119]), and others have failed to replicate (e.g., [113,120]). Such challenges call into question—or, minimally, complicate—the idea that language understanding is always and necessarily grounded in sensorimotor representations. Revising theoretical perspectives on embodied language critically requires more clarity and consensus about the implicated domains and the conditions under which activation of non-linguistic representations occurs. By offering numerous testable predictions and making explicit the standards of neuroscientific evidence required to support the exportation hypothesis (\textbf{Box 4}), we hope to move research on language understanding forward in productive way. Relatedly, additional theoretical progress will require a concerted effort to link findings across studies and fields. The literature on embodied language so far makes little contact with studies of the same extra-linguistic systems in other settings, potentially exacerbated by the pervasive use of group averaging approaches. Functional localizers—which we emphasize in our proposal—ensure that the ‘same’ (i.e., functionally-equivalent) brain region is identified across individuals and studies, allowing us to more directly connect language activation in extra-linguistic brain regions to prior studies of those same regions in other contexts. Such links will be essential for uncovering the precise contributions of these areas to language understanding.
\end{framed}
\begin{center}
  \centering 
  \captionof{figure}{\small\textbf{Information in language is exported to non-linguistic brain systems for deep understanding.} \textbf{A)} The language system (red) exports information to extra-linguistic systems that support deep understanding of linguistic content. Candidate exportation destinations include regions specialized for thinking about the contents of someone else’s or one’s own mind [26], regions specialized for scene understanding and navigation [22,28], regions that construct mental models of the physical world [31], regions that represent and process amodal semantic information [33], perceptual and motor regions (e.g., [21–23]), and regions that integrate information into a coherent representation of the situation that is being described [90,115]. These exportation destination regions are visualized with broad masks delineating where the relevant functional regions are found; although these masks may overlap, the functional regions within individuals are distinct. \textbf{B)} The language system supports shallow language understanding. Shallow language understanding entails recognizing individual words and how they go together within the structure of the sentence (i.e., parsing the surface form of the sentence), and then constructing a compositional representation of the sentence’s meaning, perhaps represented in a high-dimensional vector space. Importantly, these shallow representations are only defined with respect to other linguistic elements and patterns of their use that we have encountered during our lifetime of experience with language, making no contact with other aspects of our lived experience. \textbf{C)} Extra-linguistic systems support deep language understanding. They can be engaged both by linguistic and nonlinguistic inputs, which is why we do not consider them part of the language system. During language understanding these regions may augment the shallow representations constructed by the core language system with perceptual or motor features, translate them into a format compatible with probabilistic programing engines or symbolic problem solvers, or transform them in some other way.}
  \includegraphics[scale=0.33]{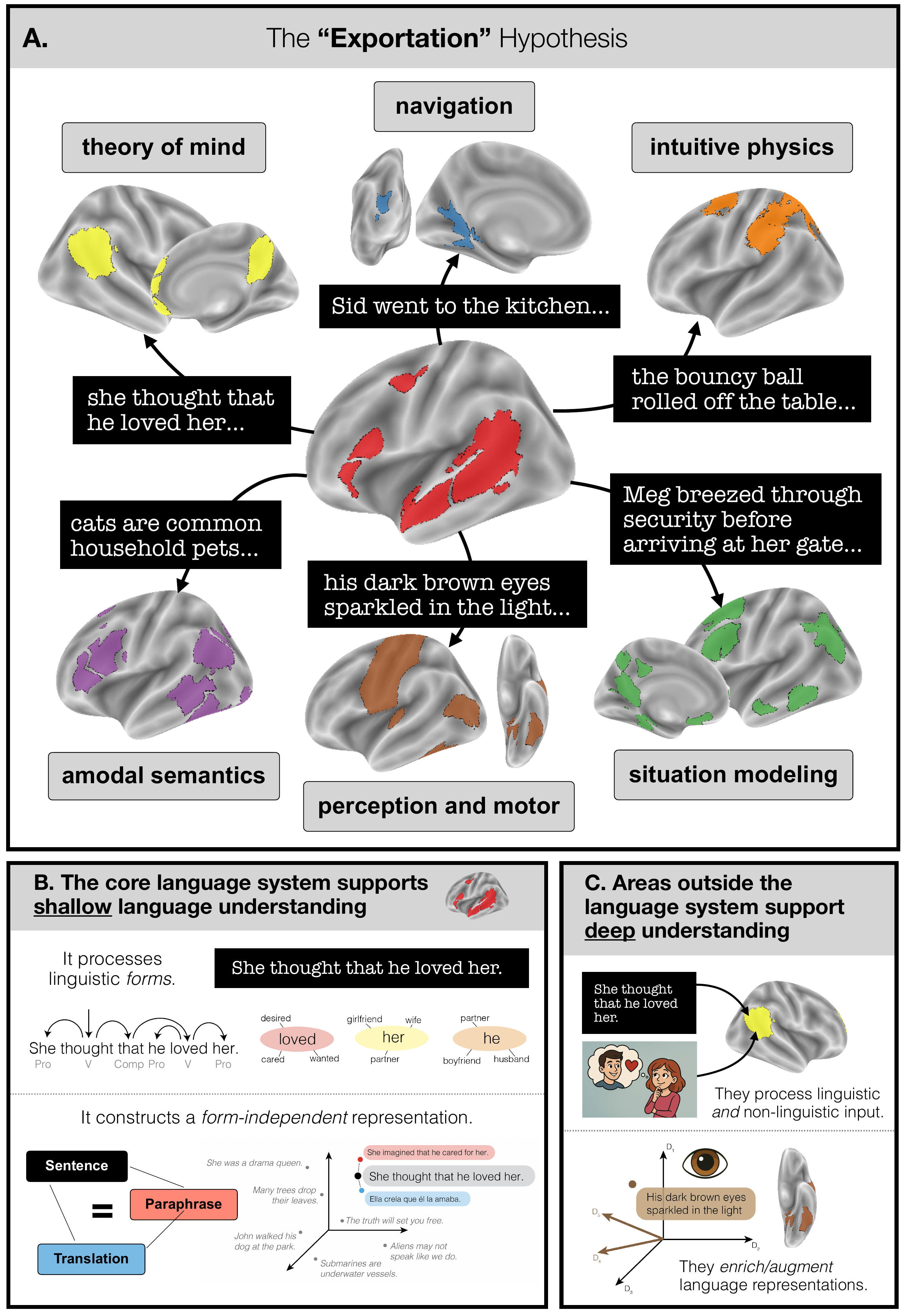}
  \clearpage 
  \label{fig:my_image}
\end{center}
\begin{spacing}{1.25}
This emerging picture of the functional architecture of the human brain [37] empowers us to take a new look at longstanding questions about language comprehension. In particular, we consider three key questions. First, how deeply is information processed within the core language system (Section 1)? Second, which brain regions outside the language system additionally support language understanding, and to what extent (Section 2)? And third, when does exportation happen, and how might it be implemented (Section 3)?

Before tackling these questions, we make three clarifications. First, our hypothesis is distinct from the claim that the whole brain—not a single, specialized network—is responsible for language processing (\textbf{Box 2}). Brain regions that, we hypothesize, language information is exported to are not language-specific: they can be strongly engaged by non-linguistic inputs (\textbf{Figure 1C}). These systems are not part of the language system simply because they can be accessed through language, any more than they are part of the visual system because they can be engaged through vision. The core language system, however, is selective for language. Second, our argument is not definitional: we are not arguing that certain mental processes should or should not “count” as language. Rather, we are advocating that language understanding—while necessarily engaging core linguistic processing—may engage additional mental processes that are not linguistic in nature (e.g., reasoning about other minds, objects, and places; accessing our knowledge of the world; etc.). Our proposal does not depend on whether these additional processes are considered part of ‘language’ or ‘cognition’ more broadly. 

Third, our critical question is not whether exportation \textit{can} happen—information conveyed in language is likely capable of penetrating even the most remote corners of our mind under the right circumstances, such as when we are asked a question or given an instruction (e.g., \textit{What is the appropriate tip for an eighty-dollar restaurant bill? What is the most efficient route from work to home? Imagine your mother’s face.}). Instead, our question concerns the division of labor between the core language system and other brain systems in generating the mental representations we construct when understanding language. We aim to determine the circumstances under which brain systems beyond the language network are recruited and to characterize their contribution to language understanding in cognitive, neural, and computational terms. \end{spacing}
\vspace{-\baselineskip}
\begin{framed}
\textbf{Box 2: A unifying account of language processing in the brain.}

Many past studies have reported neural responses in cortical regions outside the classic language areas when people process words, sentences, or stories. For example, brain responses to words in a narrative can be well predicted across the cortex by high-dimensional linguistic embeddings, such as representations from LLMs [75,76,121]. Other studies have reported stronger and more consistent responses (across individuals) in non-language-specific brain regions to a coherent verbal narrative than to a scrambled version where sentences don’t go together [70,88,89,91]. Finally, various studies have found that some brain regions outside of the core language system are semantically tuned to particular types of linguistic content [66,69–71,75,76,122–124]. Taken together, these and related findings demonstrate that regions outside of the core language areas can be engaged by language, and that some of these regions may exhibit a preference for certain content. These findings have created a fundamental tension in language neuroscience between studies that have argued for the existence of a language-specialized network (e.g., [20]), and those that have argued that “the entire brain is at work” when we process language (e.g., [125,126]).

However, these two sets of empirical findings are not at odds. In fact, many of the effects reported above (e.g., the preference of some brain areas for certain content) would be expected under the exportation hypothesis. However, we argue for two fundamental differences between the contributions of the core language system and the many other brain regions that can be engaged during language understanding. First, all linguistic input (spoken, written, or signed) is necessarily processed by the core language system initially. Further processing by downstream systems occurs only after an initial representation has been constructed by the core language system, and may sometimes not happen at all. This temporal staging of initial linguistic processing and subsequent simulation-based or conceptual processing resembles prior proposals (e.g., [10]). Second, the various areas outside the core language system are not language-specific: although they \textit{can} be engaged by linguistic inputs, they can also be engaged by visual and other non-verbal inputs (\textbf{Figure 1C}), which is why we do not consider these regions part of the language system. This perspective reconciles the sometimes-broad activation of many brain regions in response to language with studies reporting a subset of these regions to be strikingly language-specific (e.g., [20]).
\end{framed}

\section*{How deeply is information processed within the core language system?}
\label{sec:headings}
\begin{spacing}{1.25}
fMRI research over the last 15 years has identified and characterized the brain's core language system with new precision (see [20] for review). Consisting of temporal and frontal left-hemisphere regions, this system responds when we listen to or read language or when we speak, but not when we engage in apparently similar but nonlinguistic processes, such as arithmetic, music perception, understanding gestures, or reasoning (for reviews, see [20,38]). Damage to these regions leads to linguistic deficits [39]. All regions of the language network show sensitivity to structure at the sublexical, word, and sentence level [20], in line with usage-based accounts of language processing (e.g., [40]), which draw no sharp distinctions by unit size (sounds, words, sentences). 

During comprehension, the language system’s goal is to extract information from word sequences, which requires recognizing familiar words and larger constructions and figuring out how they go together within the sentence (\textbf{Figure 1B}). The end product is a representation that captures both structural and co-occurrence-based information encoded in a sentence. Several lines of evidence suggest that this representation is quite \textit{abstract}, no longer tied to the particular words. For example, representations appear to be similar across sentence paraphrases and across languages [41,42] (\textbf{Figure 1B}). Furthermore, the representations constructed within the language system share some features with representations extracted from meaningful visual inputs: pictures and silent videos elicit some response in the language areas, although substantially lower than verbal stimuli [43], and representations from neural network vision models can be used to predict responses to sentences in these areas [44,45]. Presumably, this similarity between language- and vision-derived representations (and likely other representations derived from perceptual and motor experiences in the world) stems from the fact that both language and visual experiences capture the structure of the world [46,47].

Critically, although abstract, we argue that these meaning-approximating representations are shallow: they are derived solely from our knowledge of language statistics, akin to representations in early versions of large language models trained on only text (\textbf{Box 3}; note that we are not arguing that all of language is acquired through co-occurrence-based learning). Even severe damage to the brain’s language areas can leave conceptual processing intact, suggesting that whatever representations and computations this system supports cannot be critical to world knowledge and reasoning [43,48]. Although much of our knowledge of the world (e.g., factual information) is learned through language, that need not mean this information is stored long-term in our language system. Consistent with this idea, the representations in the language system lack sensitivity to real-world plausibility (or relatedly, factual accuracy) of a sentence, as evidenced by similarly strong responses to plausible sentences such as \textit{“Sally went to the park to walk her dog”} and well-formed but nonsensical sentences like the famous example \textit{“Colorless green ideas sleep furiously”} [49] (\textbf{Figure 1B}). These nonsensical sentences abide by the rules of English syntax, but do not express a plausible meaning; the fact that the language areas respond to them just as strongly as to plausible sentences suggests that the interpretations constructed within the language system are not constrained by our knowledge of the world. This feature is of course “by design”: although language statistics capture many aspects of our world knowledge (e.g., [50]), we can use language to tell each other anything, including things that are false, nonsensical, and surprising. (\textit{Did you know that armadillos are actually bulletproof, or that owls don’t have eyeballs?}). How would we learn new things about the world through language, if the system were not flexible enough to build representations for concepts and ideas that we have not encountered before?

Taken in tandem, the available evidence suggests that the core language system is only capable of \textbf{shallow understanding}: it doesn’t really “understand” things, it only knows how people talk about them. This is not to say that the representations constructed by the language system are not useful or powerful. As noted above, language statistics reflect many aspects of the structure of the world, and—as we have learned from LLMs—training on language statistics alone is sufficient to produce grammatically well-formed and semantically coherent discourse, answer questions, and carry out simple conversations [51,52] (see \textbf{Box 3}). In fact, these shallow representations may carry enough information about the world for many of our day-to-day activities, like exchanging pleasantries with the receptionist at the doctor’s office or half-listening to one’s spouse complain about their day.

However, even if these shallow approximations of meaning are sometimes sufficient, when we read or listen to language, our minds often do more. We construct mental models of what the language is describing and rich representations of meaning that connect to our broader knowledge of the world, which suggests that the language system is often not the end point of linguistic processing (see [20] and [53] for related discussions), much as the visual system is not the end point of visual processing.
\end{spacing}
\begin{framed}
\textbf{Box 3: Using LLMs to better understand shallow and deep language understanding.}

Throughout this piece we have argued that the brain’s language system constructs a shallow representation of meaning. But what would a system that has only processed language shallowly really \textit{understand}? The clearest examples of such a system are early versions of modern large language models (LLMs) like GPT-2 [51]. These systems exhibit remarkable mastery of linguistic rules and patterns—which we termed “formal competence” in a recent piece [53]—akin to the brain’s core language system, yet these early models struggle with using language to do things in the world (i.e., “functional competence” [53]). However, the precise boundary between formal and functional competence is somewhat elusive, especially with respect to meaning. Linguistic patterns capture a wealth of knowledge about the world [46,47,50,127], but consensus is still lacking about the kinds of meanings that can, in principle, be captured by language statistics alone vs. those that require a deeper level of processing (e.g., see [128] for evidence that language models without grounding struggle to represent sensorimotor features, cf. [129]); nor is there consensus about the amount or kinds of language training data needed, in practice, for artificial systems to mimic human-like shallow understanding. Insofar as small LLMs resemble the brain’s core language regions [130,131] and capture aspects of human linguistic behavior (e.g., reading times [132] or linguistic judgements [133]), we can use them to gain insights into what shallow language understanding—without access to the more sophisticated knowledge and cognitive machinery—entails.

In addition to using LLMs to understand shallow representations of meaning, larger contemporary models may help understand the exportation process. The artificial systems that most directly parallel the architecture of the human brain are LLMs augmented with various extralinguistic components: vision models, physics and theory of mind engines, formal logic provers, enhanced memory mechanisms, and many more (e.g., [54–56,134]). However, even LLMs trained on only text, without additional built-in nonlinguistic modules (e.g. [135]) can spontaneously develop similar brain-like functional specializations for language and non-linguistic aspects of cognition [136,137]. These brain-like AI systems can be used to understand the exportation process mechanistically, including developing and evaluating computationally precise hypotheses about whether/when exportation will occur, whether exported information is broadcasted widely or routed to specific downstream targets, and how linguistic representations are modified when exported. 
\end{framed}
\section*{Which brain regions outside the language system additionally support language understanding?}
\label{sec:headings}
\begin{spacing}{1.25}
If the core language system does not fully process the meaning of a linguistic input, what other systems could? In this section, we discuss some of the brain regions that may support deep language understanding (\textbf{Figure 1C}). At a computational level, these brain systems may, for example, augment the shallow representations constructed by the core language system with perceptual or motor features [14], translate them into a format compatible with probabilistic programming engines [54,55] or symbolic problem solvers [56], or transform them in some other way. Whether these downstream representations are continuous/distributional [57] or symbolic and propositional [58] is an important question, but none of our key claims critically depend on their particular representational formats. As discussed in the Introduction, our key claims also do not critically depend on whether the mental processes supported by these brain systems are considered part of language or cognition more broadly.

Below we summarize available evidence for exportation with an emphasis on studies that satisfy our methodological desiderata (\textbf{Box 4}); for many brain areas, however, direct evidence of exportation is limited, so we focus on the \textit{predictions} of the exportation hypothesis.
\end{spacing}
\begin{framed}
\textbf{Box 4: What would count as evidence of extra-linguistic engagement?}

What would strong evidence for the exportation hypothesis look like?  We advocate three criteria:
\begin{enumerate}
    \item \textit{\textbf{Interpreting responses is easiest when the brain region in question is functionally-specific.}}  Inferences of the engagement of specific mental processes from activation in particular anatomical locations (or "reverse inference" [138]) are strongest for brain areas that are functionally-specific. For example, engagement of the FFA during language understanding would provide strong evidence that a face percept has been constructed, whereas engagement of anterior cingulate cortex would be consistent with numerous disparate cognitive processes.
    \item \textit{\textbf{The brain region should be functionally identified within individual participants.}} Because the anatomical location of functional regions varies across participants, standard group analyses blur the responses of each region with its anatomical neighbors. Thus, a precise measure of the engagement of a given functionally-specific brain region requires first identifying that region within individual brains with a standard ‘localizer’ task. For example, if we want to test the role of the PPA in understanding navigation-relevant language content, we should identify it in each participant with a contrast of viewing scenes versus objects. For many systems—such as the language network—one can also identify them using functional correlations derived from resting state (e.g., [139]) or task-based (e.g., [140]) timeseries. 
    \item \textit{\textbf{The non-language-specific system should exhibit an above-baseline response to language.}} There are various measures for assessing whether a brain region is engaged in, or contains information that is relevant to, language understanding. The strongest evidence for such engagement would be reliable, above-baseline responses to language that are similar in magnitude to responses elicited by other stimuli and tasks in that region. It is challenging to argue that a region that is not engaged above baseline is performing a critically important computation during language understanding, even if the region exhibits a reliable preference for certain types of linguistic content (e.g., a region that \textit{deactivates less} to navigation-related content than social content). This said, we do not rule out the importance—perhaps even critical role—of language-relevant information encoded in extra-linguistic brain regions that does not result in average above-baseline activation
\end{enumerate}
\end{framed}
\begin{spacing}{1.25}
\subsection*{\textit{a. Brain Regions that Build Mental Models of Minds, Objects, and Places}}

\paragraph{Theory of mind.} There is at least one case where the evidence is clear that information is exported from the language system to be further processed elsewhere: theory of mind (ToM), the ability to construct a mental model of someone else’s (or one’s own) mind. Extensive studies by Saxe and colleagues have demonstrated the highly specific fMRI response of an area in the right temporo-parietal junction (rTPJ) when participants think about the content of thoughts and beliefs [26,59]. This region is functionally correlated with a set of other regions (e.g., [59,60]) that together constitute the ToM network [26–28].

In many studies of the ToM network, participants are asked to read a short passage and then choose the correct completion of the story, which requires inferring the beliefs of a character in the passage. The ToM network responds strongly during this task (and the response occurs during the part of the passage that talks about mental states), but not during a logically similar task in which the passage requires an inference about a physical representation [26]. The ToM network responds even if no question is asked and the participant need only read and understand the text (e.g., [26,61]), and it is also engaged when viewing silent movies that require thinking about mental states [62,63]. Finally, language and ToM areas are partially synchronized during story comprehension (e.g., [63]), which provides indirect evidence of the exportation process as well. Theory of mind is therefore a case where a full understanding of information delivered through language entails exporting representations from the brain's language system to another functionally-specific region, the rTPJ, along with the broader ToM network.

\paragraph{Intuitive physics.} Just as we build mental models of each other’s minds, we also build mental models of the physical world to understand what we are looking at, predict what will happen next, and plan actions [66]. Growing evidence implicates regions of the parietal and frontal cortex in this intuitive physical reasoning (aka the "Physics Network" [31,64]). Although these regions have been primarily studied using nonverbal visual scenarios, they have also been shown to respond more strongly when solving verbal problems about physical (compared to social) scenarios [65,66], suggesting that information from the language system can be exported to the Physics Network for further processing. Regions in the frontal and parietal lobes have also been shown to respond to language describing actions (e.g., [67–71]), however, it is not known whether the implicated regions comprise, or are simply adjacent to, the Physics Network given the use of group averaging approaches in prior studies (\textbf{Box 4}).

\paragraph{Navigation and scene understanding.} Another kind of mental model concerns the map of space we use to navigate in the world. Extensive work in human cognitive neuroscience has identified key brain regions underlying our navigational ability (see [29] for review), including the parahippocampal place area (PPA, [22,72]), the occipital place area (OPA, [72]), and retrosplenial cortex (RSC, or “medial place area”, MPA [22,73]). Recent evidence suggests that these three scene-processing areas border the Default Network (DN), specifically DN-A [34], which is preferentially connected to the anterior hippocampus [74] and has been argued to represent and process place-related information [28]. Consistent with the exportation hypothesis, preliminary evidence indicates that these regions can be engaged by understanding paragraphs that contain information about places and navigation [75,76]. Given what is known about dissociations among the different place areas based on vision research, the exportation hypothesis also specifically predicts that during language understanding the PPA would become engaged by e.g., a vivid verbal description of the shape of a room [22], the OPA by a description of an open doorway [77], and the RSC when listening to someone’s experience of regaining their bearings after emerging from a subway [78]. 

\subsection*{\textit{b. Brain Regions that House Perceptual and Motor Representations}}

\paragraph{Motor.}The proposal we advance here resembles many classic ideas from embodied and grounded cognition [14,16], which have primarily focused on the relationship between linguistic and sensorimotor representations. In a hallmark study, Hauk et al. [79] reported that passive reading of words describing actions of the mouth (“lick”), arm (“pick”) or leg (“kick”) activated regions of motor cortex overlapping with or adjacent to those activated by corresponding actual movements, and numerous subsequent studies have shown activations in primary motor or premotor cortex to language (e.g., [68–71,80,81]; see [82,83] for review). Several of these studies faced methodological limitations (\textbf{Box 4}) and substantial disagreement remains on the conditions under which motor representations are activated (e.g., [84]), and on the necessity and sufficiency of these activations for language understanding (e.g., [18,85]). Nonetheless, these and other studies of grounded cognition highlight the potentially important role of sensorimotor brain areas in language understanding (\textbf{Box 1}).

\paragraph{Perception.} An engaging novel or story can conjure up vivid mental imagery. According to the exportation hypothesis, this is possible due to exporting information to corresponding brain areas. Consider visual processing. Some of the strongest evidence for functional specificity in the cortex comes from studies of mid-level and high-level vision, where cortical regions respond selectively to mid-level visual features like motion and to high-level categories like faces, places, and bodies [21–23]. It is well established that these regions can be activated when participants are explicitly instructed to mentally imagine faces or places [73] or motion [86]. But it is not known whether these functionally-specific visual regions are engaged when understanding verbal descriptions of visual scenes. Would your FFA fire up for you to really understand a passage like this: \textit{"He was not over thirty. His eyes were very dark brown and there was a hint of brown pigment in his eyeballs. His cheek bones were high and wide, and strong deep lines cut down his cheeks, in curves beside his mouth. His upper lip was long, and since his teeth protruded, the lips stretched to cover them, for this man kept his lips closed.”} (Grapes of Wrath, Steinbeck).

\subsection*{\textit{c. Other Brain Regions}}

In addition to the areas outlined above, there are likely many other brain areas that could support aspects of language understanding. For example, language routinely taps into our general conceptual knowledge about the world, drawing on our amodal semantic system [17,33]. The semantic regions may also serve to facilitate or mediate interactions between the language system and the brain areas discussed above (see [18] for discussion). Similarly, many have argued that the goal of language understanding is to construct "situation models" that combine the kind of mental models discussed above with our general world knowledge into a single coherent representation of what we are hearing or reading (i.e., \textit{Who did what, where?}) [2,4,7,9]. Some studies have shown that the Default Network [87], unlike the language system, integrates information over long contexts for both verbal and visual inputs, potentially constructing these evolving situation-level representations [88–91]. However, recent evidence suggests that the DN (as originally construed; e.g., [87]) consists of two parallel, subnetworks: the Theory of Mind system and Default Network-A (DN-A) [27,34], both of which we have introduced earlier. These networks have similar overall anatomy but are clearly dissociable within individual participants. Therefore, some of the early findings implicating the DN in situation model construction should be revisited to more critically assess the nature of its contributions to language understanding, including testing whether the situation models we construct from language arise from a dynamic—and possibly hippocampal-dependent [92–94]—coordination of representations constructed by multiple content-specific systems, rather than a single brain network integrating information across domains. Finally, there may be other regions not discussed here (e.g., regions that process emotion [62,95,96]) that contribute in important ways to deep language understanding.
\end{spacing}

\section*{When does exportation happen, and how might it be implemented?}
\label{sec:headings}
\begin{spacing}{1.25}
In addition to discovering the full ontology of extra-linguistic mechanisms that may get engaged during language comprehension, another important research goal is to understand what factors affect the likelihood of information exportation. We briefly highlight three possible factors. First, longer verbal excerpts may be more likely to be exported. Given the language system is memory-limited and only processes short segments of language [91,97,98], input that extends beyond a sentence or so must be passed to other systems that can construct temporally extended contexts (such as the DN-A and the Theory of Mind networks [91]). By exporting constructed meanings to other systems, the language system frees up space for new incoming input [99]. Second, traits, states, and goals of the comprehender plausibly matter, including how much experience they have with the topic (e.g., [100,101]), how alert they are (e.g., [102]), how able they are to engage in perceptual imagery (e.g., [103]), how interested they are in the content (e.g., [104]), and whether or not the comprehender has to perform an explicit task (e.g., [65]). Imagine, for example, listening to a doctor explain treatments for a particular condition before vs. after you’ve been diagnosed with that condition: the likelihood of exportation in the latter case is potentially much higher. The same is likely to hold when reading a compelling narrative about a protagonist whose fate you have become vicariously invested in (e.g., see [70]). Indeed, previous behavioral work has shown that people employ different mental processes (e.g., perceptual and motor simulation, mentalizing) to differing degrees during natural reading [105], which may be attributable, in part, to individuals having different goals when reading the same text. Third, exportation critically depends on the language system being able to extract from the surface form a representation that is abstract enough to interface with downstream systems. Consequently, the difficulty of extracting such representations due to the complexity of the form (e.g., infrequent words or complex constructions), time pressure, or insufficient proficiency of the comprehender (e.g., a young child or a non-native speaker) may make exportation less likely.

When exportation does take place, how is this process implemented? One possibility is that the language system preferentially sends information to the most likely target destination (i.e., \textbf{“routing”}). Under this account, the representation constructed by the language system is routed to the correct downstream system based on some features of the linguistic representation. A different possibility is that representations constructed by our language system are sent to all possible downstream systems all the time (i.e., \textbf{“broadcasting”}). Under this account, information from the language system is broadly diffused, and the exportation destination—not the language system—evaluates the representational content of the input to determine whether to process the information further (see [106] for additional discussion of the neural processes that enable communication between brain networks). Distinguishing these possibilities would require spatiotemporally sensitive neural measures with broad coverage, which no single method currently available in human cognitive neuroscience provides. Therefore, making progress towards a mechanistic understanding of the exportation process will likely require the coordinated use of multiple techniques or \textit{in silico} experimentation in artificial systems (e.g., [107], see \textbf{Box 3}). We also emphasize that the flow of information is likely bidirectional. During production, the language system must receive information from brain regions that construct the meaning to be communicated. And even during comprehension, substantial behavioral evidence indicates that language processing is affected by extra-linguistic information sources, including the visual environment (e.g., [108]), non-verbal signals (e.g., gestures [109]), prior linguistic context (e.g., [110]), world knowledge (e.g., [111]), and knowledge of what information is accessible to our interlocutor (e.g., [112], see \textit{\textbf{Outstanding questions}}).
\end{spacing}
\begin{framed}
\textbf{Outstanding questions}
\begin{enumerate}
    \item How much of everyday language understanding takes place within the core language system? Is exportation the exception or the rule?
    \item How are language representations "translated" into a format that other brain regions can process?
    \item When information delivered through language is not exported, what has the listener understood?
    \item What training data, architectures, and learning algorithms are needed for artificial systems to acquire human-like shallow understanding? Are extra-linguistic modules necessary for deep understanding in these systems?
    \item What specifically is represented in each of the exportation destinations and how is this representation formatted?
    \item What stimulus- and person-factors affect exportation? 
    \item Does the language system route information to the appropriate extra-linguistic system, or broadcast it broadly across the brain?
    \item How is exportation from the language system coordinated with information flow in the opposite direction (from extra-linguistic brain areas to the language system)?    
\end{enumerate}
\end{framed}
\newpage
\section*{Concluding remarks}
\label{sec:headings}
\begin{spacing}{1.25}
In this opinion piece, we have proposed that deep language understanding requires exporting information from the brain's core language system to other systems that build mental models of the minds, objects, and places described in the linguistic input, that store our memories and world knowledge, and that contain our perceptual and motor representations. Some evidence for this idea already exists, but our goal here is to outline a \textit{conceptual framework} for thinking about what the brain does with linguistic input beyond the parsing of sentences and construction of shallow approximations of linguistic meaning carried out in the language network and to outline many \textit{testable hypotheses}.

Several caveats should be noted. First, strong empirical evidence for the exportation hypothesis is so far restricted to very few systems, and some studies find no evidence for exportation (e.g., [113,114]). Although suggestive evidence for the exportation hypothesis abounds (see \textbf{Box 1} and \textbf{2}), methodological limitations (e.g., reliance on group-averaging approaches, \textbf{Box 1} and \textbf{4}) restrict the conclusions we can draw from much past work. Second, to achieve deep understanding, we have argued that activation of an extra-linguistic brain system during language processing is \textit{necessary}. However, clear fMRI evidence of this activation does not imply that deep understanding was achieved (i.e., that activation alone may not be \textit{sufficient}). For example, think of trying, but failing, to construct a mental model from a linguistic description (e.g., because the description is not clear enough, or because you are distracted while trying to do it). Third, it is possible that some, but not all, of the systems outlined above will be routinely recruited during language understanding. Such a result would shed light on the kinds of meanings that we, in practice, approximate using our linguistic representations vs. those that require deeper conceptual or simulation-based processing to construct. Finally, there is currently no consensus on how to behaviorally evaluate the depth of language understanding; yet such measures would be critical for falsifying the idea that deep understanding requires exportation. The field of education research, which is currently grappling with how to probe students’ understanding in the age of AI, may eventually provide some solutions.

In addition to advancing our understanding of the human brain, our hope is that this framework will also inform cognitive theories and the development of more human-like artificial agents. As we discuss in the Introduction, a large body of work in cognitive psychology has demonstrated that we \textit{can} construct mental situation models when understanding language [2,4,9]. However, neural evidence may help us to identify circumstances when we get by with shallow understanding only. Similarly, neuroscience may provide insights into the format of these situation models and may help to characterize the timing and process of updating this model during online language comprehension. On the AI side, considerable interest exists in developing LLM architectures that can learn from text more efficiently and that can perform in more robust and human-like ways across cognitive tasks (\textbf{Box 3}). A better understanding of the division of labor among the core language system and other brain networks, as well as their interactions, may ultimately enable the development of new artificial systems that reflect the inductive biases present in the human brain. 

By speculating about the brain regions that may work in concert with the language system, the circumstances under which they may be recruited, and the methods available to test these ideas, we have proposed a broad research program that harnesses recent progress in human cognitive neuroscience to advance a richer theory of what language understanding entails cognitively, neurally, and computationally.
\end{spacing}
\newpage
\begin{framed}
\textbf{Glossary}
\begin{itemize}
    \item \textbf {Broadcasting:} Another hypothesis for how information is exported to downstream systems, whereby information is sent from the core language system to all possible downstream systems, which then evaluate whether to process the information further.
    \item \textbf{Core language system:} A network of frontal and temporal brain regions that are highly specialized for language processing, strongly functionally and anatomically interconnected, and causally important in comprehending and producing language. The language system builds shallow representations of word- and sentence-level meanings. 
    \item \textbf{Deep understanding:} Building rich mental models of the people, objects, places, and situations described in a language input. These models may incorporate our world knowledge, autobiographical memories, and/or perceptual and motor representations. 
    \item \textbf{Exportation:} The process by which information delivered through language is transferred out of our core language system to one or more other brain regions for further processing.
    \item \textbf{Routing:} One hypothesis for how information is exported from the language system to downstream systems, whereby information is selectively directed (e.g., by the language system or a mediating brain region) to the most appropriate downstream system. 
    \item \textbf{Shallow understanding:} Building a representation of linguistic input based solely on knowledge of patterns of language use. This process requires recognizing familiar elements (e.g., morphemes, words, larger constructions), figuring out how those elements relate to one another, and ultimately combining them into a form-independent abstract representation (i.e. one that generalizes across paraphrases and languages) that can interface with downstream systems. Importantly, the representations constructed during shallow understanding are grounded only to the language statistics, and not tied to our lived experience or broader knowledge of the world.
    \item \textbf{Situation model:} A representation of the ‘situation’ that is being described by a language input, including the characters and objects described, their environments and spatial locations, as well the goals of the individuals, and the temporal and causal relations among events. 
\end{itemize}
\end{framed}

\section*{Acknowledgments}
\label{sec:headings}
We thank Andrea de Varda, Ted Gibson, Steve Piantadosi, Rebecca Saxe, Thomas Clark, and members of the Kanwisher lab for helpful discussion and comments on the manuscript. We also thank Emalie McMahon for help with the figure. CC was supported by the Kempner Institute for the Study of Natural and Artificial Intelligence at Harvard University. EF was partially supported by U01 award NS121471 from NINDS; EF and NK were partially supported by funds from the Simons Foundation awarded to the Simons Center for the Social Brain at MIT, from the McGovern Institute for Brain Research, and from the Siegel Family Quest for Intelligence.

\newpage
\section*{References}
\begin{enumerate}
    \item Zhang, Y. et al. (2025) Comparative illusions are evidence of rational inference in language comprehension. J. Exp. Psychol. Gen. 154, 2752–2771
    \item Bransford, J.D. et al. (1972) Sentence memory: A constructive versus interpretive approach. Cogn. Psychol. 3, 193–209
    \item Bower, G.H. and Morrow, D.G. (1990) Mental Models in Narrative Comprehension. Science (1979). 247, 44–48
    \item Johnson-Laird, P.N. (1983) Mental Models: Towards a Cognitive Science of Language, Inference and Consciousness., Harvard University Press
    \item Gernsbacher, M.A. (1990) Language comprehension as structure building, Erlbaum
    \item Graesser, A.C. et al. (1994) Constructing inferences during narrative text comprehension. Psychol. Rev. 101, 371–395
    \item Zwaan, R.A. and Radvansky, G.A. (1998) Situation Models in Language Comprehension and Memory. Psychol. Bull. 123, 162–185
    \item Jackendoff, R.S. and Erk, K.E. (2025) Toward a Deeper Lexical Semantics. Top. Cogn. Sci. 17, 962–972
    \item van Dijk, T.A. and Kintsch, W. (1983) Strategies in discourse comprehension, Academic Press
    \item Barsalou, L.W. et al. (2008) Language and simulation in conceptual processing. In Symbols and Embodiment: Debates on Meaning and Cognition (de Vega, M., ed), pp. 245–284, Oxford University Press
    \item Louwerse, M. and Jeuniaux, P. (2008) Language comprehension is both embodied and symbolic. In Symbols and Embodiment: Debates on Meaning and Cognition (de Vega, M., ed), pp. 309–326, Oxford University Press
    \item Pavio, A. (1971) Imagery and verbal processes, Holt, Rinehart, and Winston
    \item Caramazza, A. et al. (1990) The multiple semantics hypothesis: Multiple confusions? Cogn. Neuropsychol. 7, 161–189
    \item Barsalou, L.W. (2008) Grounded cognition. Annu. Rev. Psychol. 59, 617–645
    \item Louwerse, M. and Jeuniaux, P. (2010) The linguistic and embodied nature of conceptual processing. Cognition 114, 96–104
    \item Barsalou, L.W. et al. (2003) Grounding conceptual knowledge in modality-specific systems. Trends Cogn. Sci. 7, 84–91
    \item Lambon Ralph, M.A. et al. (2016) The neural and computational bases of semantic cognition. Nat. Rev. Neurosci. 18, 42–55
    \item Mahon, B.Z. and Caramazza, A. (2008) A critical look at the embodied cognition hypothesis and a new proposal for grounding conceptual content. Journal of Physiology-Paris 102, 59–70
    \item Meteyard, L. et al. (2012) Coming of age: A review of embodiment and the neuroscience of semantics. Cortex 48, 788–804
    \item Fedorenko, E. et al. (2024) The language network as a natural kind within the broader landscape of the human brain. Nat. Rev. Neurosci. 25, 289–312
    \item Kanwisher, N. et al. (1997) The Fusiform Face Area: A Module in Human Extrastriate Cortex Specialized for Face Perception. J. Neurosci. 17, 4302–4311
    \item Epstein, R. and Kanwisher, N. (1998) A cortical representation of the local visual environment. Nature 392, 598–601
    \item Downing, P.E. et al. (2001) A cortical area selective for visual processing of the human body. Science (1979). 293, 2470–2473
    \item Norman-Haignere, S. et al. (2015) Distinct Cortical Pathways for Music and Speech Revealed by Hypothesis-Free Voxel Decomposition. Neuron 88, 1281–1296
    \item Overath, T. et al. (2015) The cortical analysis of speech-specific temporal structure revealed by responses to sound quilts. Nat. Neurosci. 18, 903–911
    \item Saxe, R. and Kanwisher, N. (2003) People thinking about thinking people: The role of the temporo-parietal junction in “theory of mind.” Neuroimage 19, 1835–1842
    \item DiNicola, L.M. et al. (2020) Parallel distributed networks dissociate episodic and social functions within the individual. J. Neurophysiol. 123, 1144–1179
    \item Deen, B. and Freiwald, W.A. (2025) Parallel systems for social and spatial cognition reaching the cortical apex. Proc. Natl. Acad. Sci. U. S. A. 122, e2520067122
    \item Epstein, R.A. and Baker, C.I. (2019) Scene Perception in the Human Brain. Annu. Rev. Vis. Sci. 5, 373–397
    \item Culham, J.C. and Valyear, K.F. (2006) Human parietal cortex in action. Curr. Opin. Neurobiol. 16, 205–212
    \item Fischer, J. et al. (2016) Functional neuroanatomy of intuitive physical inference. Proc. Natl. Acad. Sci. U. S. A. 113, E5072–E5081
    \item Isik, L. et al. (2017) Perceiving social interactions in the posterior superior temporal sulcus. Proc. Natl. Acad. Sci. U. S. A. 114, E9145–E9152
    \item Ivanova, A.A. et al. (2025) Semantic reasoning takes place largely outside the language network. bioRxiv DOI: 10.64898/2025.12.07.692873
    \item Braga, R.M. and Buckner, R.L. (2017) Parallel Interdigitated Distributed Networks within the Individual Estimated by Intrinsic Functional Connectivity. Neuron 95, 457-471.e5
    \item Duncan, J. (2010) The multiple-demand (MD) system of the primate brain: mental programs for intelligent behaviour. Trends Cogn. Sci. 14, 172–179
    \item Kean, H. et al. (2026) Evidence from formal logical reasoning reveals that the language of thought is not natural language. Proceedings of the National Academy of Sciences 123, e2520095123
    \item Kanwisher, N. (2025) Animal models of the human brain: Successes, limitations, and alternatives. Curr. Opin. Neurobiol. 90, 102969
    \item Fedorenko, E. et al. (2024) Language is primarily a tool for communication rather than thought. Nature 630, 575–586
    \item Fridriksson, J. et al. (2018) Anatomy of aphasia revisited. Brain 141, 848–862
    \item Bybee, J. (2010) Language, Usage and Cognition, Cambridge University Press
    \item Kauf, C. et al. (2024) Lexical semantic content, not syntactic structure, is the main contributor to ANN-brain similarity of fMRI responses in the language network. Neurobiology of Language 5, 7–42
    \item de Varda, A.G. et al. (2026) Multilingual Computational Models Capture a Shared Meaning Component in Brain Responses across 21 Languages. bioRxiv DOI: 10.1101/2025.02.01.636044
    \item Ivanova, A.A. et al. (2021) The Language Network Is Recruited but Not Required for Nonverbal Event Semantics. Neurobiology of Language 2, 176–201
    \item Saha, S. et al. (2025) Modeling the language cortex with form-independent and enriched representations of sentence meaning reveals remarkable semantic abstractness. ArXiv DOI: https://doi.org/10.48550/arXiv.2510.02354
    \item Small, H. et al. (2025) Ubiquitous cortical sensitivity to visual information during naturalistic, audiovisual movie viewing. PsyArxiv
    \item Huh, M. et al. (2024) The Platonic Representation Hypothesis. in Proceedings of the 41st International Conference on Machine Learning (PMLR) 235, pp. 20617–20642
    \item de Varda, A.G. et al. (2025) A distributional model of concepts grounded in the spatial organization of objects. J. Mem. Lang. 142, 104624
    \item Dickey, M.W. and Warren, T. (2015) The influence of event-related knowledge on verb-argument processing in aphasia. Neuropsychologia 67, 63–81
    \item Kauf, C. et al. (2024) Linguistic inputs must be syntactically parsable to fully engage the language network. bioRxiv DOI: 10.1101/2024.06.21.599332
    \item Grand, G. et al. (2022) Semantic projection recovers rich human knowledge of multiple object features from word embeddings. Nat. Hum. Behav. 6, 975–987
    \item Radford, A. et al. (2019) Language Models are Unsupervised Multitask Learners. in arXiv
    \item Brown, T.B. et al. (2020) Language models are few-shot learners. in arXiv
    \item Mahowald, K. et al. (2024) Dissociating language and thought in large language models. Trends Cogn. Sci. 28, 517–540
    \item Zhang, C. et al. (2023) Grounded physical language understanding with probabilistic programs and simulated worlds. in Proceedings of the Annual Meeting of the Cognitive Science Society
    \item Wong, L. et al. (2023) From Word Models to World Models: Translating from Natural Language to the Probabilistic Language of Thought. ArXiv DOI: https://doi.org/10.48550/arXiv.2306.12672
    \item Olausson, T.X. et al. (2023) LINC: A Neurosymbolic Approach for Logical Reasoning by Combining Language Models with First-Order Logic Provers. in Conference on Empirical Methods in Natural Language Processing, pp. 5153–5176
    \item Piantadosi, S.T. et al. (2024) Why concepts are (probably) vectors. Trends Cogn. Sci. 28, 844–856
    \item Piccinini, G. (2025) Neural Hardware for the Language of Thought: New Rules for an Old Game. ArXiv DOI: https://doi.org/10.48550/arXiv.2510.10251
    \item Saxe, R. and Wexler, A. (2005) Making sense of another mind: The role of the right temporo-parietal junction. Neuropsychologia 43, 1391–1399
    \item Saxe, R. and Powell, L.J. (2006) It’s the thought that counts: specific brain regions for one component of theory of mind. Psychol. Sci. 17, 692–699
    \item Bruneau, E.G. et al. (2012) Distinct roles of the ‘Shared Pain’ and ‘Theory of Mind’ networks in processing others’ emotional suffering. Neuropsychologia 50, 219–231
    \item Jacoby, N. et al. (2016) Localizing Pain Matrix and Theory of Mind networks with both verbal and non-verbal stimuli. Neuroimage 126, 39–48
    \item Paunov, A.M. et al. (2019) Functionally distinct language and theory of mind networks are synchronized at rest and during language comprehension. J. Neurophysiol. 121, 1244–1265
    \item Pramod, R.T. et al. (2025) Decoding predicted future states from the brain’s “physics engine.” Sci. Adv. 11, 7429
    \item Pramod, R.T. et al. (2024) A Region in Human Left Prefrontal Cortex Selectively Engaged in Causal Reasoning. in Proceedings of the Annual Meeting of the Cognitive Science Society, 46
    \item Jack, A.I. et al. (2013) fMRI reveals reciprocal inhibition between social and physical cognitive domains. Neuroimage 66, 385–401
    \item Aflalo, T. et al. (2020) A shared neural substrate for action verbs and observed actions in human posterior parietal cortex. Sci. Adv. 6
    \item Tettamanti, M. et al. (2005) Listening to action-related sentences activates fronto-parietal motor circuits. J. Cogn. Neurosci. 17, 273–281
    \item Chow, H.M. et al. (2014) Embodied comprehension of stories: interactions between language regions and modality-specific neural systems. J. Cogn. Neurosci. 26, 279–295
    \item Kurby, C.A. and Zacks, J.M. (2013) The activation of modality-specific representations during discourse processing. Brain Lang. 126, 338–349
    \item Raposo, A. et al. (2009) Modulation of motor and premotor cortices by actions, action words and action sentences. Neuropsychologia 47, 388–396
    \item Nakamura, K. et al. (2000) Functional delineation of the human occipitotemporal areas related to face and scene processing a PET study. Brain 123, 1903–1912
    \item O’Craven, K.M. and Kanwisher, N. (2000) Mental imagery of faces and places activates corresponding stiimulus-specific brain regions. J. Cogn. Neurosci. 12, 1013–1023
    \item Braga, R.M. (2025) The canonical default network comprises parallel distributed networks with distinct medial temporal lobe connections. Curr. Opin. Behav. Sci. 66, 101610
    \item Antonello, R. et al. (2024) Generative causal testing to bridge data-driven models and scientific theories in language neuroscience. ArXiv DOI: https://doi.org/10.48550/arXiv.2410.00812
    \item Singh, C. et al. (2025) Evaluating scientific theories as predictive models in language neuroscience. bioRxiv DOI: 10.1101/2025.08.12.669958
    \item Kamps, F.S. et al. (2016) The occipital place area represents the local elements of scenes. Neuroimage 132, 417–424
    \item Baumann, O. and Mattingley, J.B. (2010) Medial parietal cortex encodes perceived heading direction in humans. J. Neurosci. 30, 12897–12901
    \item Hauk, O. et al. (2004) Somatotopic Representation of Action Words in Human Motor and Premotor Cortex. Neuron 41, 301–307
    \item Willems, R.M. et al. (2010) Body-specific representations of action verbs: neural evidence from right- and left-handers. Psychol. Sci. 21, 67–74
    \item Aziz-Zadeh, L. et al. (2006) Congruent Embodied Representations for Visually Presented Actions and Linguistic Phrases Describing Actions. Current Biology 16, 1818–1823
    \item Pulvermüller, F. (2005) Brain mechanisms linking language and action. Nat. Rev. Neurosci. 6, 576–582
    \item Fischer, M.H. and Zwaan, R.A. (2008) Embodied Language: A Review of the Role of the Motor System in Language Comprehension. Quarterly Journal of Experimental Psychology 61, 825–850
    \item Schuil, K.D.I. et al. (2013) Sentential context modulates the involvement of the motor cortex in action language processing: An fMRI study. Front. Hum. Neurosci. 7, 41161
    \item Zwaan, R.A. (2014) Embodiment and language comprehension: reframing the discussion. Trends Cogn. Sci. 18, 229–234
    \item Goebel, R. et al. (1998) The constructive nature of vision: direct evidence from functional magnetic resonance imaging studies of apparent motion and motion imagery. Eur. J. Neurosci. 10, 1563–1573
    \item Buckner, R.L. et al. (2008) The brain’s default network: anatomy, function, and relevance to disease. Ann. N. Y. Acad. Sci. 1124, 1–38
    \item Xu, J. et al. (2005) Language in context: Emergent features of word, sentence, and narrative comprehension. Neuroimage 25, 1002–1015
    \item Yarkoni, T. et al. (2008) Neural substrates of narrative comprehension and memory. Neuroimage 41, 1408–1425
    \item Baldassano, C. et al. (2017) Discovering Event Structure in Continuous Narrative Perception and Memory. Neuron 95, 709-721.e5
    \item Lerner, Y. et al. (2011) Topographic mapping of a hierarchy of temporal receptive windows using a narrated story. J. Neurosci. 31, 2906–2915
    \item Duff, M.C. and Brown-Schmidt, S. (2012) The hippocampus and the flexible use and processing of language. Front. Hum. Neurosci. 6
    \item Milivojevic, B. et al. (2016) Coding of Event Nodes and Narrative Context in the Hippocampus. J. Neurosci. 36, 12412–12424
    \item Dijksterhuis, D.E. et al. (2024) Pronouns reactivate conceptual representations in human hippocampal neurons. Science (1979). 385, 1478–1484
    \item Ferstl, E.C. et al. (2005) Emotional and temporal aspects of situation model processing during text comprehension: an event-related fMRI study. J. Cogn. Neurosci. 17, 724–739
    \item Hsu, C.T. et al. (2015) The emotion potential of words and passages in reading Harry Potter – An fMRI study. Brain Lang. 142, 96–114
    \item Blank, I.A. and Fedorenko, E. (2020) No evidence for differences among language regions in their temporal receptive windows. Neuroimage 219, 116925
    \item Jacoby, N. and Fedorenko, E. (2020) Discourse-level comprehension engages medial frontal Theory of Mind brain regions even for expository texts. Lang. Cogn. Neurosci. 35, 780–796
    \item Christiansen, M.H. and Chater, N. (2015) The Now-or-Never bottleneck: A fundamental constraint on language. Behavioral and Brain Sciences 39
    \item Amalric, M. and Dehaene, S. (2016) Origins of the brain networks for advanced mathematics in expert mathematicians. Proc. Natl. Acad. Sci. U. S. A. 113, 4909–4917
    \item Beilock, S.L. et al. (2008) Sports experience changes the neural processing of action language. Proc. Natl. Acad. Sci. U. S. A. 105, 13269–13273
    \item Cohen, L. et al. (2021) How does inattention affect written and spoken language processing? Cortex 138, 212–227
    \item Zeman, A. (2024) Aphantasia and hyperphantasia: exploring imagery vividness extremes. Trends Cogn. Sci. 28, 467–480
    \item Olson, H.A. et al. (2024) Personalized neuroimaging reveals the impact of children’s interests on language processing in the brain. Imaging Neuroscience 2, 1–14
    \item Mak, M. and Willems, R.M. (2019) Mental simulation during literary reading: Individual differences revealed with eye-tracking. Lang. Cogn. Neurosci. 34, 511–535
    \item Seguin, C. et al. (2023) Brain network communication: concepts, models and applications. Nat. Rev. Neurosci. 24, 557–574
    \item Jain, S. et al. (2024) Computational Language Modeling and the Promise of in Silico Experimentation. Neurobiology of Language 5, 80–106
    \item Tanenhaus, M.K. et al. (1995) Integration of visual and linguistic information in spoken language comprehension. Science (1979). 268, 1632–1634
    \item Holle, H. and Gunter, T.C. (2007) The role of iconic gestures in speech disambiguation: ERP evidence. J. Cogn. Neurosci. 19, 1175–1192
    \item Nieuwland, M.S. and Van Berkum, J.J.A. (2006) When Peanuts Fall in Love: N400 Evidence for the Power of Discourse. J. Cogn. Neurosci. 18, 1098–1111
    \item Hagoort, P. et al. (2004) Integration of Word Meaning and World Knowledge in Language Comprehension. Science (1979). 304, 438–441
    \item Hanna, J.E. et al. (2003) The effects of common ground and perspective on domains of referential interpretation. J. Mem. Lang. 49, 43–61
    \item Dravida, S. et al. (2013) People can understand descriptions of motion without activating visual motion brain regions. Front. Psychol. 4, 39942
    \item Bedny, M. et al. (2008) Concepts Are More than Percepts: The Case of Action Verbs. J. Neurosci. 28, 11347
    \item Radvansky, G.A. and Zacks, J.M. (2014) Event Perception. In Event Cognition, pp. 40–56, Oxford University Press
    \item Glenberg, A.M. and Kaschak, M.P. (2002) Grounding language in action. Psychon. Bull. Rev. 9, 558–565
    \item Stanfield, R.A. and Zwaan, R.A. (2001) The effect of implied orientation derived from verbal context on picture recognition. Psychol. Sci. 12, 153–156
    \item Pecher, D. et al. (2003) Verifying different-modality properties for concepts produces switching costs. Psychol. Sci. 14, 119–124
    \item Körner, A. et al. (2023) Embodied Processing at Six Linguistic Granularity Levels: A Consensus Paper. J. Cogn. 6, 60–60
    \item Zwaan, R.A. and Pecher, D. (2012) Revisiting Mental Simulation in Language Comprehension: Six Replication Attempts. PLoS One 7, e51382
    \item Huth, A.G. et al. (2016) Natural speech reveals the semantic maps that tile human cerebral cortex. Nature 532, 453–458
    \item Ferstl, E.C. and von Cramon, D.Y. (2007) Time, space and emotion: fMRI reveals content-specific activation during text comprehension. Neurosci. Lett. 427, 159–164
    \item Whitney, C. et al. (2009) Neural correlates of narrative shifts during auditory story comprehension. Neuroimage 47, 360–366
    \item Speer, N.K. et al. (2009) Reading Stories Activates Neural Representations of Visual and Motor Experiences. Psychol. Sci. 20, 989
    \item Drijvers, L. et al. (2025) Language is widely distributed throughout the brain. Nat. Rev. Neurosci. 26, 189–189
    \item Aliko, S. et al. (2025) The entire brain, more or less, is at work: ‘Language regions’ are artefacts of averaging. bioRxiv DOI: 10.1101/2023.09.01.555886
    \item Rinaldi, L. and Marelli, M. (2020) Maps and Space Are Entangled with Language Experience. Trends Cogn. Sci. 24, 853–855
    \item Xu, Q. et al. (2025) Large language models without grounding recover non-sensorimotor but not sensorimotor features of human concepts. Nat. Hum. Behav. 9, 1871–1886
    \item Marjieh, R. et al. (2024) Large language models predict human sensory judgments across six modalities. Sci. Rep. 14, 21445-
    \item Hosseini, E.A. et al. (2024) Artificial Neural Network Language Models Predict Human Brain Responses to Language Even After a Developmentally Realistic Amount of Training. Neurobiology of Language 5, 43–63
    \item AlKhamissi, B. et al. (2025) From Language to Cognition: How LLMs Outgrow the Human Language Network. in Proceedings of the Conference on Empirical Methods in Natural Language Processing (EMNLP), pp. 24321–24339
    \item Oh, B.D. and Linzen, T. (2026) To model human linguistic prediction, make LLMs less superhuman. Trends Cogn. Sci. DOI: 10.1016/j.tics.2026.05.008
    \item Hu, J. et al. (2024) Language models align with human judgments on key grammatical constructions. Proc. Natl. Acad. Sci. U. S. A. 121, e2400917121
    \item Dong, C. V. et al. (2025) Towards large language models with human-like episodic memory. Trends Cogn. Sci. 29, 928–941
    \item Yang, A. et al. (2024) Qwen2.5 Technical Report. ArXiv DOI: https://doi.org/10.48550/arXiv.2412.15115
    \item AlKhamissi, B. et al. (2024) The LLM Language Network: A Neuroscientific Approach for Identifying Causally Task-Relevant Units. in Association for Computational Linguistics, 1, pp. 10887–10911
    \item Han, P. et al. (2026) Modular Cognitive Architecture Emerges in Large Language Models. ArXiv
    \item Poldrack, R.A. (2011) Inferring Mental States from Neuroimaging Data: From Reverse Inference to Large-Scale Decoding. Neuron 72, 692–697
    \item Braga, R.M. et al. (2020) Situating the left-lateralized language network in the broader organization of multiple specialized large-scale distributed networks. J Neurophysiol 124, 1415–1448
    \item Shain, C. and Fedorenko, E. (2026) A language network in the individualized functional connectomes of 1199 human brains doing arbitrary tasks. Nat. Commun. 17, 8180-
\end{enumerate}
\end{document}